\documentclass[conference]{IEEEtran}
\IEEEoverridecommandlockouts

\usepackage{cite}
\usepackage{amsmath,amssymb,amsfonts}
\usepackage{graphicx}
\usepackage{textcomp}
\usepackage{xcolor}
\usepackage{algorithmic}
\usepackage[ruled,linesnumbered]{algorithm2e}
\usepackage{enumerate}
\usepackage{url}
\usepackage{multirow}
\usepackage{booktabs}
\usepackage{threeparttable}
\usepackage{endnotes}
\usepackage{bbding}
\usepackage{CJKutf8}
\usepackage{bm}

\def\BibTeX{{\rm B\kern-.05em{\sc i\kern-.025em b}\kern-.08em
    T\kern-.1667em\lower.7ex\hbox{E}\kern-.125emX}}
\begin{document}

\begin{CJK*}{UTF8}{gbsn}

\title{Automatic Severity Classification of Coronary Artery Disease via Recurrent Capsule Network}
% \title{Recurrent Capsule Network for Relation Extraction: A Practical Application to the Severity Classification of Coronary Artery Disease}
\author{\IEEEauthorblockN{Qi Wang$^1$, Jiahui Qiu$^1$, Yangming Zhou$^{1}$, Tong Ruan$^{1,*}$, Daqi Gao$^1$ and Ju Gao$^{2,*}$}
\IEEEauthorblockA{$^1$School of Information Science and Engineering, East China University of Science and Technology, Shanghai 200237, China\\
$^2$Shanghai Shuguang Hospital, Shanghai 200021, China \\
$^*$Corresponding authors\\
Emails: \{ruantong@ecust.edu.cn, gaoju@smmail.cn\}
}
}

\maketitle

\begin{abstract}

Coronary artery disease (CAD) is one of the leading causes of cardiovascular disease deaths. CAD condition progresses rapidly, if not diagnosed and treated at an early stage may eventually lead to an irreversible state of the heart muscle death. Invasive coronary arteriography is the gold standard technique for CAD diagnosis. Coronary arteriography texts describe which part has stenosis and how much stenosis is in details. It is crucial to conduct the severity classification of CAD. In this paper, we employ a recurrent capsule network (RCN) to extract semantic relations between clinical named entities in Chinese coronary arteriography texts, through which we can automatically find out the maximal stenosis for each lumen to inference how severe CAD is according to the improved method of Gensini. Experimental results on the corpus collected from Shanghai Shuguang Hospital show that our proposed method achieves an accuracy of 97.0\% in the severity classification of CAD.

\end{abstract}

\begin{IEEEkeywords}
Electronic health records, coronary artery disease, severity classification; recurrent capsule network; relation extraction
\end{IEEEkeywords}

% ------------------------------------------------------------------------------------------
\section{Introduction}
\label{Sec:Introduction}

A large amount of electronic health records (EHRs) data has been accumulated since the wide use of medical information systems in China. However, most of these records are written in natural language, which cannot be processed by computers directly. For example, coronary arteriography is the gold standard technique for the diagnosis of coronary artery disease (CAD). However, after writing the results of coronary angiography on EHRs, doctors have to classify the severity of CAD manually based on the coronary angiography results, according to the method of Gensini~\cite{gensini1983more}. Since coronary arteriography texts describes which part has stenosis and how much stenosis is in details, if we can extract the relations between clinical named entities in coronary arteriography texts, we can automatically find out the maximal stenosis for each lumen, and inference how severe CAD is under the guide of the method of Gensini. Thus, one of the key issues to conduct the severity classification of CAD is relation extraction, i.e., extract the relations between clinical named entities in coronary arteriography texts.

\begin{figure}[!htb]
\begin{center}
\includegraphics[width=3.2in]{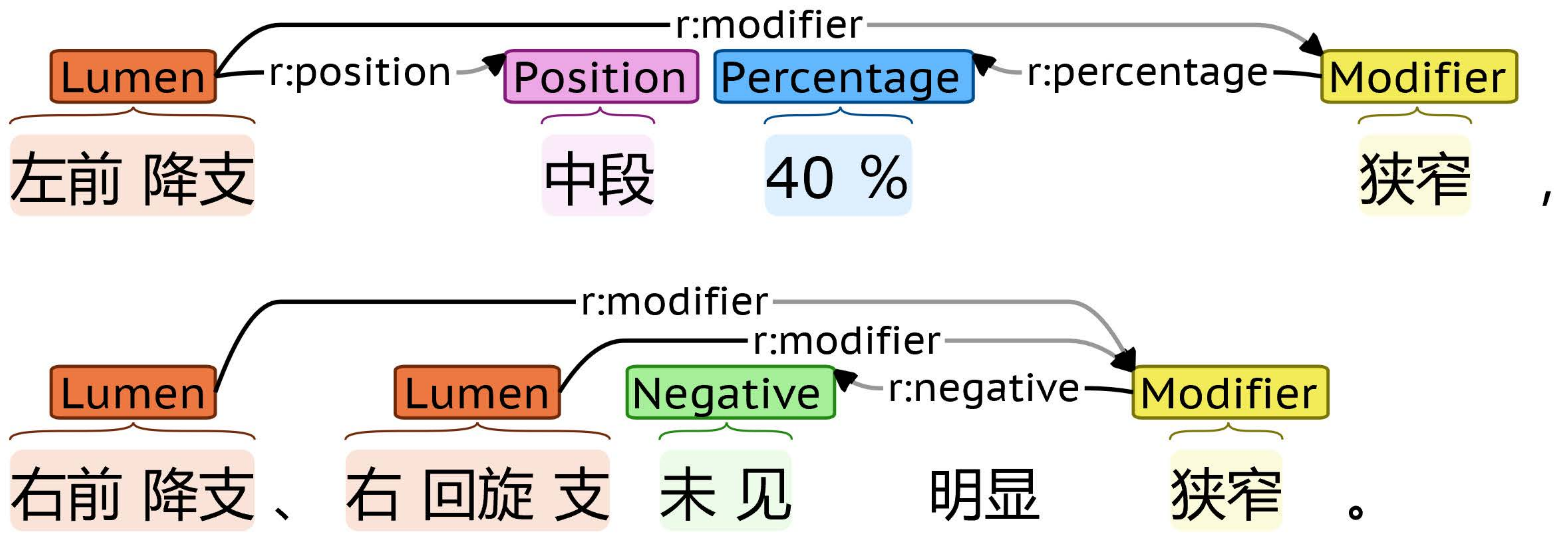}
\caption{An illustrative example of relation extraction.}
\label{Fig:relationExample}
\end{center}
\end{figure}

Fig.~\ref{Fig:relationExample} shows an illustrative example of relation extraction in the sentence ``\begin{CJK*}{UTF8}{gbsn}左前降支中段40\%狭窄，右前降支、右回旋支未见明显狭窄。\end{CJK*}'' (40\% of the middle of left anterior descending branch has stenosis and no obvious stenosis has been seen in the right anterior descending branch and the right circumflex coronary artery), in which the relations
$<$\begin{CJK*}{UTF8}{gbsn}左前降支\end{CJK*}, r:modifier(e1,e2), \begin{CJK*}{UTF8}{gbsn}狭窄\end{CJK*}$>$ ($<$left anterior descending branch, r:modifier(e1,e2), stenosis$>$)
, $<$\begin{CJK*}{UTF8}{gbsn}左前降支\end{CJK*}, r:position(e1,e2), \begin{CJK*}{UTF8}{gbsn}中段\end{CJK*}$>$ ($<$left anterior descending branch, r:position(e1,e2), middle$>$)
, $<$40\%, r:percentage(e2,e1), \begin{CJK*}{UTF8}{gbsn}狭窄\end{CJK*}$>$ ($<$40\%, r:percentage(e2,e1), stenosis$>$)
, $<$\begin{CJK*}{UTF8}{gbsn}右前降肢\end{CJK*}, r:modifier(e1,e2), \begin{CJK*}{UTF8}{gbsn}狭窄\end{CJK*}$>$ ($<$right anterior descending branch, r:modifier(e1,e2), stenosis$>$)
, $<$\begin{CJK*}{UTF8}{gbsn}右回旋支\end{CJK*}, r:modifier(e1,e2), \begin{CJK*}{UTF8}{gbsn}狭窄\end{CJK*}$>$ ($<$right circumflex coronary artery, r:modifier(e1,e2), stenosis$>$)
and $<$\begin{CJK*}{UTF8}{gbsn}未见\end{CJK*}, r:negative(e2,e1), \begin{CJK*}{UTF8}{gbsn}狭窄\end{CJK*}$>$ ($<$not seen, r:negative(e2,e1), stenosis$>$) should be extracted.

The relation extraction task is full of challenges due to the following reasons: (1) Different modifiers may share the same lumen, and different lumina may share the same modifiers. It denotes long-term dependency in a coronary arteriography sentence. (2) The same sentence can be expressed in different ways, such that ``\begin{CJK*}{UTF8}{gbsn}40\%狭窄\end{CJK*}'' and ``\begin{CJK*}{UTF8}{gbsn}狭窄40\%\end{CJK*}'' express the same meaning (i.e. 40\% of sth. has stenosis).

A number of methods have been proposed for the relation extraction task. These methods are usually based on supervised methods or semi-supervised methods via deep neural networks, such as convolutional neural networks (CNNs)~\cite{zeng2014relation,nguyen2015relation,zeng2015distant} and recurrent neural networks (RNNs)~\cite{zhang2015relation,zhou2016attention,xu2015classifying}. However, these existing methods all use only a neuron (i.e. scalar) to represent the classification probability via a sigmoid function or a softmax function, which limit the expressive ability of neural networks.

In this paper, we employ a recurrent capsule network (RCN) model to extract entity relations. Specifically, after clinical named entity recognition (CNER), words and their corresponding entity type features are first transferred into embedding vectors, then fed into a recurrent layer to capture high-level features. Finally, a capsule layer is used for relation classification, where the length of the capsules (i.e. vectors) is used to represent the probability that the corresponding relation exists, and different orientation of a capsule can represent different cases under the relation, so the model can achieve a stronger expressive ability. Extensive experimental results on the coronary arteriography texts collected from Shanghai Shuguang Hospital show that our RCN model obtains the best performance compared with baseline methods.

The contributions of this paper can be summarized as follows.
\begin{itemize}
    \item We present an effective method for the automatic severity classification of CAD from EHRs. A recurrent capsule network model is employed to extract semantic relations between clinical named entities in Chinese texts.
    \item We conduct extensive experiments on the coronary arteriography texts collected from Shanghai Shuguang Hospital. Experimental results demonstrate that our proposed method outperforms the baseline methods.
\end{itemize}

The rest of the paper is organized as follows. In Section \ref{Sec:Related Works}, we briefly review the related work on relation extraction. In Section \ref{Sec:Severity Classification of CAD}, we present an effective method for the automatic severity classification of CAD. We report the computational results in Section \ref{Sec:Experimental Studies}. Section \ref{Sec:Analysis and Discussion} is devoted to experimental analysis and some discussions. Finally, conclusions and possible research directions are given in Section \ref{Sec:Conclusions and Future Works}.

\section{Related Works}
\label{Sec:Related Works}

Due to its practical significance, relation extraction has attracted considerable research effort in the last decades and a lot of methods have been proposed in the literature. The existing methods can be roughly classified into three categories, namely supervised methods, semi-supervised methods and joint extraction methods.

Traditionally, supervised methods consider the relation extraction task as a relation classification problem, and utilize statistical machine learning methods to address it~\cite{miller2000novel,culotta2006integrating}. Typical methods are support vector machines (SVMs)~\cite{zelenko2003kernel,zhao2005extracting}. However, these statistical methods rely on pre-defined features, which makes their development costly. What's more, feature engineering, i.e. finding the best set of features for classification, is more of an art than a science, incurring extensive trial-and-error experiments. Lately, with the popularity of deep learning, most focus has shifted towards deep neural networks. Deep learning method is an end-to-end solution, which can learn features from a training set automatically, so it is not necessary to deign hand-crafted features, and biomedical knowledge resources are also not prerequisite. Socher et al.~\cite{socher2012semantic} employed a recursive neural network model that learns compositional vector representations for phrases  and sentences of arbitrary syntactic type and length. Zeng et al.~\cite{zeng2014relation} and Nguyen and Grishman~\cite{nguyen2015relation} utilized the convolutional neural networks (CNNs) to extract lexical and sentence level features. Zhang and Wang~\cite{zhang2015relation} used a bidirectional recurrent neural network (Bi-RNN) with max-pooling to classify relations. Zhou et al.~\cite{zhou2016attention} exploited an attention-based bidirectional long short-term memory (Bi-LSTM) network for relation classification. Xu et al.~\cite{xu2015classifying} classified relations via LSTM networks along shortest dependency paths. Raj et al.~\cite{raj2017learning} utilized a convolutional recurrent neural network (CRNN) architecture that combines RNNs and CNNs in sequence. They also evaluated an attention-based pooling technique compared with conventional max-pooling strategies. Ren et al.~\cite{ren2018neural} introduced external text descriptions of named entities to enhance the relation classification. Christopoulou et al.~\cite{christopoulou2018walk} proposed a graphical method to extract relations.

Compared to supervised methods which requires lots of annotated corpora as training data, semi-supervised methods, also known as distant supervision, automatically annotate training data through existing knowledge bases. This type of methods assumes that if two entities have a relation in a known knowledge base, at least one sentence that contain those two entities might express that relation. Thus, in this method, an already existing knowledge base is heuristically aligned to texts, and the alignment results are treated as labeled data. However, a sentence that mentions two entities does not always express their relation in a knowledge base. That means any individual labeled sentence may give an incorrect cue, so how to address the wrong label problem is the key to distant supervision. Early works~\cite{mintz2009distant,riedel2010modeling,hoffmann2011knowledge, surdeanu2012multi} applied supervised models to elaborately designed features when obtained the labeled data through distant supervision. With the popularity of deep learning, Zeng et al.~\cite{zeng2015distant} employed piecewise CNNs with multi-instance learning for distant supervision. Lin et al.~\cite{lin2016neural} and Ji et al.~\cite{ji2017distant} introduced sentence-Level attention into CNNs to dynamically reduce the weights of the noisy instances. Feng et al.~\cite{feng2018reinforcement}, Zeng et al.~\cite{zeng2018large} and Qin et al.~\cite{qin2018robust} utilized reinforcement learning to deal with the noise of data. Qin et al.~\cite{qin2018dsgan} also tried adversarial learning to provide a cleaned dataset for relation classification.

The above two categories are pipeline methods which require to firstly recognize named entities then classify relations. Besides these two categories, some other methods tried to jointly extract entities and relations via graphical models~\cite{singh2013joint}, table representation~\cite{miwa2014modeling}, semi-Markov chains~\cite{li2014incremental} and deep neural networks~\cite{miwa2016end,zheng2017joint,zeng2018extracting}. We treat these methods as the third category, i.e., joint extraction methods.

\section{Severity Classification of CAD}
\label{Sec:Severity Classification of CAD}

\subsection{Framework}
\label{SubSec:Workflow}

\begin{figure}[!htb]
\begin{center}
\includegraphics[width=3.5in]{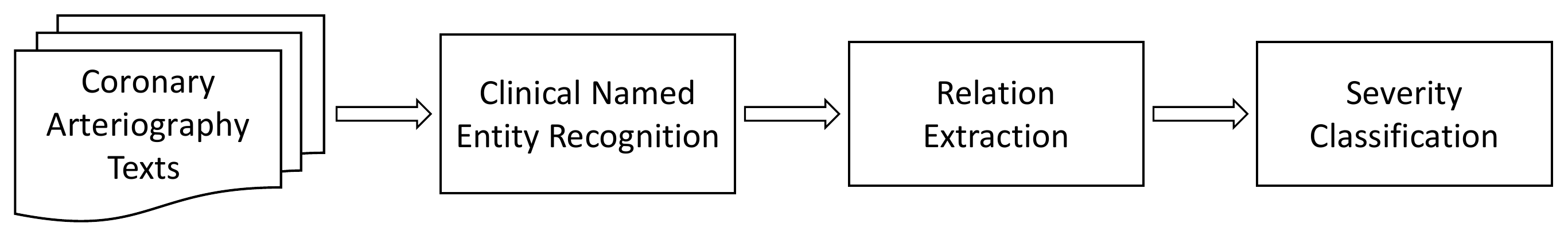}
\caption{Diagram of the severity classification of CAD.}
\label{Fig:Workflow}
\end{center}
\end{figure}

Given a coronary arteriography text, the whole diagram of the severity classification of CAD is shown in Fig.~\ref{Fig:Workflow}. Clinical named entities are first recognized and then relation extraction is made between the recognized entities. Finally, a severity score of CAD is determined for each patient based on the extracted relations according to the improved method of Gensini~\cite{gensini1983more}. Note that duo to the great challenges of extracting relations between clinical named entities, in this paper we focus on the relation extraction and employ a recurrent capsule network (RCN) model to solve it (see Section~\ref{Sec:Recurrent Capsule Network} for details).

\subsection{Clinical named entity recognition}
\label{SubSec:Clinical Named Entity Recognition}

Given a coronary arteriography text, we need to recognize some clinical named entities. These entities can be classified into following five categories.
\begin{itemize}
   \item \textbf{Lumen}: An entity that represents a body part of coronary arteries, such as ``\begin{CJK*}{UTF8}{gbsn}左主干\end{CJK*}'' (left main coronary artery), ``\begin{CJK*}{UTF8}{gbsn}左前降支\end{CJK*}'' (left anterior descending branch) and ``\begin{CJK*}{UTF8}{gbsn}左回旋支\end{CJK*}'' (left circumflex coronary artery).
   \item \textbf{Modifier}: An entity which modifies a lumen, such as ``\begin{CJK*}{UTF8}{gbsn}正常\end{CJK*}'' (normal), ``\begin{CJK*}{UTF8}{gbsn}狭窄\end{CJK*}'' (stenosis), and ``\begin{CJK*}{UTF8}{gbsn}闭塞\end{CJK*}'' (occlusion).
   \item \textbf{Negative}: An entity that indicates something does not exist, such as ``\begin{CJK*}{UTF8}{gbsn}无\end{CJK*}'' (no), ``\begin{CJK*}{UTF8}{gbsn}未\end{CJK*}'' (not), and ``\begin{CJK*}{UTF8}{gbsn}未见\end{CJK*}'' (unseen).
   \item \textbf{Position}: An entity that indicates the place where stenosis is located, such as ``\begin{CJK*}{UTF8}{gbsn}近段\end{CJK*}'' (proximal segment), ``\begin{CJK*}{UTF8}{gbsn}中段\end{CJK*}'' (middle segment), and ``\begin{CJK*}{UTF8}{gbsn}远段\end{CJK*}'' (distal segment).
   \item \textbf{Percentage}: An entity that indicates the severity of stenosis, such as ``60\%'',``\begin{CJK*}{UTF8}{gbsn}70\%\end{CJK*}'', and ``\begin{CJK*}{UTF8}{gbsn}90\%\end{CJK*}''.
\end{itemize}

Many studies have focused on the clinical named entity recognition (CNER) tasks and most of them formulate the task as a sequence labeling problem, employing various machine learning algorithms to address it~\cite{de2011machine,Habibi2017Deep}. In our previous work~\cite{wang2018incorporating}, we also proposed a CNER model which combines data-driven deep learning approaches and knowledge-driven dictionary approaches. As to this task, due to the limited entities, here we simply utilize string matching and regular matching methods for entity recognition.

\subsection{Relation extraction}
\label{SubSec:Relation Extraction}

Relation extraction is the task of finding semantic relations between pairs of entities, including modified relations, negative relations, percentage relations and position relations. It can be regarded as a multi-classification problem with two directions and an undirected no-relation class. In this paper, we employ a recurrent capsule network (RCN) model for relation extraction. We present the RCN model in Section \ref{Sec:Recurrent Capsule Network}.

\subsection{Severity classification}
\label{SubSec:Severity Classification}

Once relation extraction is finished, for each patient, a severity score of CAD is determined based on the extracted relations by using the improved method of Gensini~\cite{gensini1983more}. The scores of the right and left coronary arteries are required to obtained respectively. In all cases the angiography texts showing the most severe stenosis of each lumen is selected for grading:
\begin{equation}
score_i =
\begin{cases}
1 & \text{1\% $\leq diameter_i \leq$ 49\%}\\
2 & \text{50\% $\leq diameter_i \leq$ 74\%}\\
3 & \text{75\% $\leq diameter_i \leq$ 99\%}\\
4 & \text{$diameter_i$ = 100\%}
\end{cases}
\end{equation}
where $score_i$ is the score of the $i$-th lumen and $diameter_i$ is the maximum lesions of the $i$-th lumen's diameter. That is, lesions of $1\%$ to $49\%$ of luminal diameter are given a score of 1, those of $50\%$ to $74\%$ a score of 2, those of $75\%$ to $99\%$ a score of 3, those of 100\% (i.e. occlusion) a score of 4. Finally, for each patient, a total coronary score to reflect the extent of CAD is determined by calculating the sum of the scores for each lesion:
\begin{equation}
    score = \sum_{i} score_i
\end{equation}

We further classify the severity of CAD into three levels via the total coronary score, namely mild stenosis (between 0 and 7), moderate stenosis (between 8 and 14) and severe stenosis (over 14).

\section{Recurrent Capsule Network}
\label{Sec:Recurrent Capsule Network}

As mentioned above, our proposed method for the severity classification of CAD is composed of three components: clinical named entity recognition, relation extraction and severity classification. Since the clinical named entity recognition and the severity classification can be simply conducted by some pre-defined rules. Here we focus on the relation extraction, and employ a recurrent capsule network (RCN) to extract semantic relations between clinical named entities.

\subsection{Main architecture of RCN}
\label{SubSec: Main architecture of RCN}

\begin{figure}[!htbp]
\begin{center}
\includegraphics[width=3.2in]{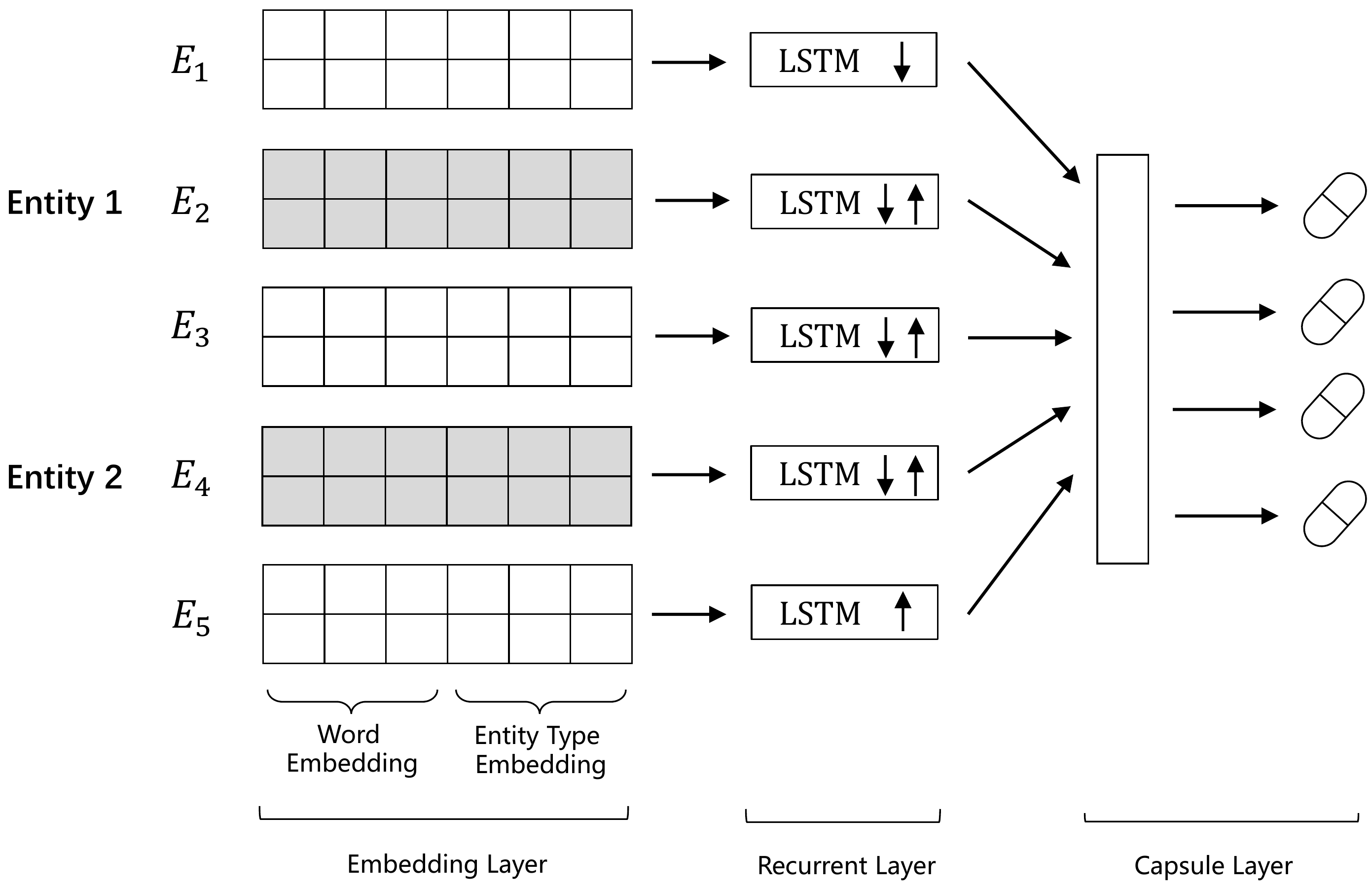}
\caption{Main architecture of the recurrent capsule network (take four classification as an example).}
\label{Fig:Architecture}
\end{center}
\end{figure}

As shown in Fig.~\ref{Fig:Architecture}, the network integrates three key components, namely embedding layer, recurrent layer and capsule layer. The Chinese words in a coronary arteriography sentence are firstly represented as distributed embedding vectors through embedding layers, and then fed into the recurrent layers. For the recurrent layer, we employ long short-term memory (LSTM)~\cite{hochreiter1997long} as the basic recurrent units to capture high-level features. Specifically, since the two entities in the sentence can divide the sentence into five segments (i.e. $E_1$, $E_2$, $E_3$, $E_4$ and $E_5$), we employ five LSTMs to handle the five segments, and output five feature vectors, respectively. Finally, an capsule layer is utilized to classify relations, where the lengths of capsules (i.e. vectors) are used to represent the probabilities whether the corresponding relation exists or not, and the orientation of a capsule can represent different cases.

\subsection{Embedding layer}
\label{SubSec:Embedding layer}

Given a coronary arteriography sentence $X={[x]}_1^T$, which is a sequence of $T$ words, the first step is to map discrete language symbols, including the words and their corresponding entity types, to distributed embedding vectors. Formally, we first lookup word embedding ${\bm{x}'}_i \in \mathbb{R}^{d_x}$ from word embedding matrix $W_x$ for each word $x_i$, where $i \in \{1,2,...,T\}$ indicates $x_i$ is the $i$-th word in $X$, and ${d_x}$ is a hyper-parameter indicating the size of word embedding. We also look up entity type embedding ${\bm{d}'}_i \in \mathbb{R}^{d_d}$ from entity type embedding matrix $W_d$ for each type of the entity which $x_i$ belongs to, where ${d_d}$ is a hyper-parameter indicating the size of entity type embedding. The final embedding vector is created by concatenating ${\bm{x}'}_i$ and ${\bm{d}'}_i$ as ${\bm{e}_i}={\bm{x}'}_i \oplus {\bm{d}'}_i$, where $\oplus$ is the concatenation operator.

\subsection{Recurrent layer}
\label{SubSec:Recurrent Layer}

The long short-term memory network (LSTM)~\cite{hochreiter1997long} is a variant of the recurrent neural network (RNN), which incorporates a gated memory-cell to capture long-range dependencies within the data and is able to avoid gradient vanishing/exploding problems caused by standard RNNs.

\begin{figure}[!hbp]
\begin{center}
\includegraphics[width=2.5in]{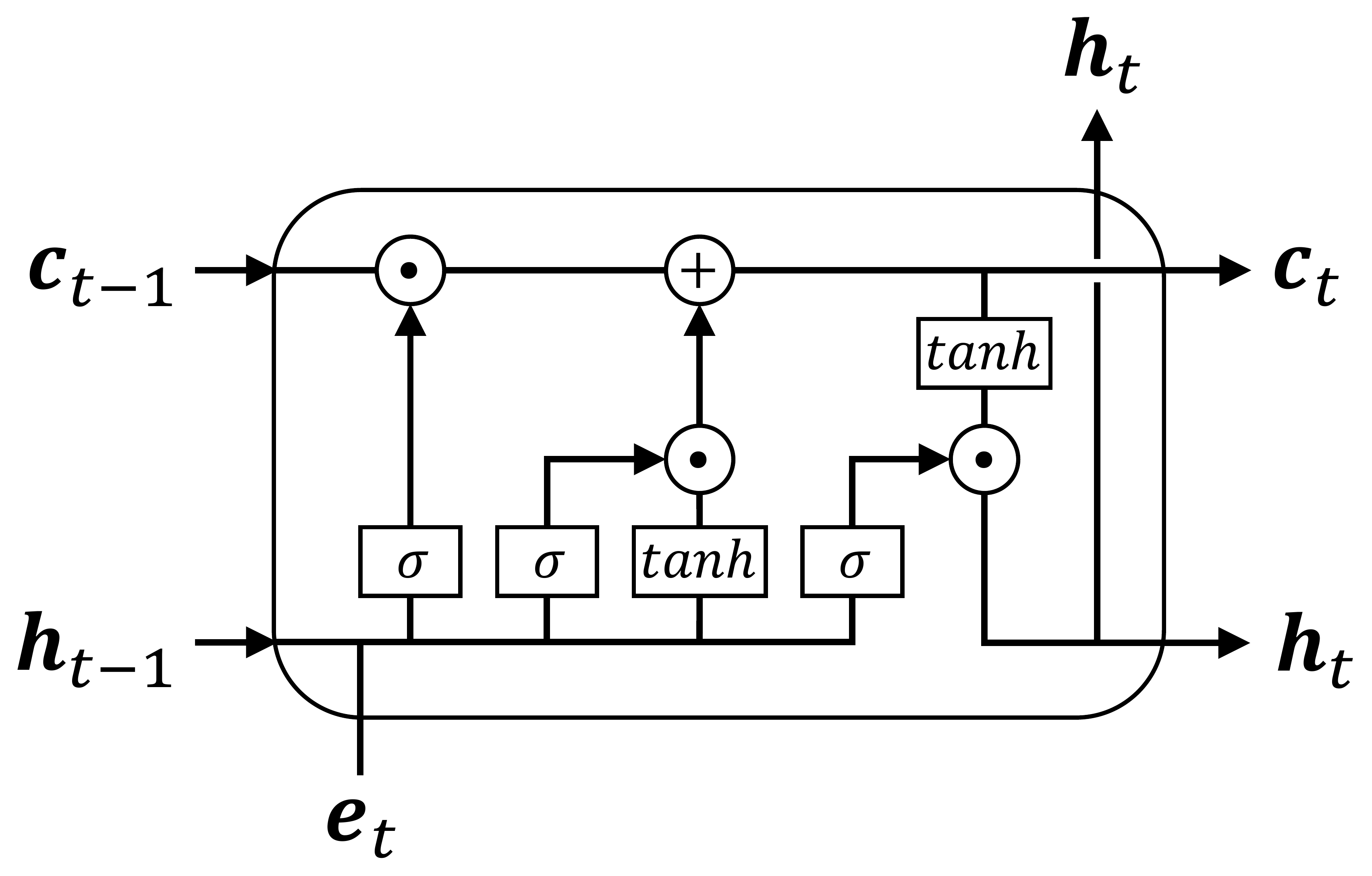}
\caption{Illustration of an LSTM cell.}
\label{Fig:LSTM_cell}
\end{center}
\end{figure}

The LSTM cell is illustrated in Fig.~\ref{Fig:LSTM_cell} For each position $t$, LSTM computes $\bm{h}_t$ with input $\bm{e}_t$ and previous state $\bm{h}_{t-1}$, as:
\begin{align}
\bm{i}_t & = \sigma(\bm{W}_i\bm{e}_t+\bm{U}_i\bm{h}_{t-1}+\bm{b}_i) \\
\bm{f}_t & = \sigma(\bm{W}_f\bm{e}_t+\bm{U}_f\bm{h}_{t-1}+\bm{b}_f) \\
\tilde{\bm{c}}_t & = \tanh(\bm{W}_c\bm{e}_t+\bm{U}_c\bm{h}_{t-1}+\bm{b}_c) \\
\bm{c}_t & = \bm{f}_t\odot \bm{c}_{t-1}+\bm{i}_t\odot \tilde{\bm{c}}_t \\
\bm{o}_t & = \sigma(\bm{W}_o\bm{e}_t+\bm{U}_o\bm{h}_{t-1}+\bm{b}_o) \\
\bm{h}_t & = \bm{o}_t\odot \tanh(\bm{c}_t)
\end{align}
where $\bm{h}$, $\bm{i}$, $\bm{f}$, $\bm{o} \in \mathbb{R}^{d_h}$ are $d_h$-dimensional hidden state (also called output vector), input gate, forget gate and output gate, respectively; $\bm{W}_i$, $\bm{W}_f$, $\bm{W}_c$, $\bm{W}_o \in \mathbb{R}^{4d_h \times d_e}$, $\bm{U}_i$, $\bm{U}_f$, $\bm{U}_c$, $\bm{U}_o \in \mathbb{R}^{4d_h\times d_h}$ and $\bm{b}_i$, $\bm{b}_f$, $\bm{b}_c$, $\bm{b}_o \in \mathbb{R}^{4d_h}$ are the parameters of the LSTM; $\sigma$ is the sigmoid function, and $\odot$ denotes element-wise production.

Due to the current state $\bm{h}_t$ also take the previous state $\bm{c}_{t-1}$ and $\bm{h}_{t-1}$ into account, the final state $\bm{h}_{T_i}$ can be thought as the representation of the whole segment $E_i$. However, the hidden state $\bm{h}_t$ of LSTM only takes information from past, not considering future information. One solution is to utilize bidirectional LSTM (Bi-LSTM)~\cite{Graves2005Framewise}, which incorporate information from both past and future. Formally, for any given sequence, the network computes both a left, $\overrightarrow{\bm{h}}_{T_i}$, and a right, $\overleftarrow{\bm{h}}_0$, representations of the sequence context at the final timestep. The representation of the whole segment $E_i$ is created by concatenating them as ${\bm{h}}=\overrightarrow{\bm{h}}_{T_i} \oplus \overleftarrow{\bm{h}}_0$.

Since the two entities in the sentence can divide the sentence into five segments, we employ five LSTMs to handle the five segments, respectively. Specifically, considering the sentences in coronary arteriography texts may very long, and the farther the distance between the context and an entity is, the less influence the context will have on  relation classification, except that Bi-LSTMs are employed for the three middle segments (i.e. $E_2$, $E_3$ and $E_4$), the left-to-right $\overrightarrow{\rm{LSTM}}$ is employed for the left-most segment (i.e. $E_1$), and the right-to-left $\overleftarrow{\rm{LSTM}}$ is employed for the right-most segment (i.e. $E_5$). Thus, the closer the context is to an entity, the more likely the context is remembered by the LSTM. .

\subsection{Capsule layer}
\label{SubSec:Capsule Layer}

The capsule layer is first proposed in ~\cite{Sabour2017Dynamic} for digit recognition. Different from traditional practice which use only a neuron to represent the classification probability via a sigmoid function or a softmax function, in this paper, a capsule is a group of neurons whose activity vector represents the instantiation parameters of a specific type of relation: The length of the activity vector is used to represent the probability that the corresponding relation exists, and different orientation of the vector can represent different cases under the relation, so the capsule can achieve a stronger expressive ability.

Considering that the length of a capsule is used as the probability of a relation, a non-linear squashing function is used to ensure that short vectors get shrunk to almost zero length and long vectors get shrunk to a length slightly below 1:
\begin{equation}
    \bm{v}_j = \frac{{||\bm{s}_j||}^2}{1+{||\bm{s}_j||}^2} \frac{\bm{s}_j}{||\bm{s}_j||}
\end{equation}
where $\bm{v}_j$ is the vector output of input capsule $\bm{s}_j$.

\begin{figure}[!htb]
\begin{center}
\includegraphics[width=3.6in]{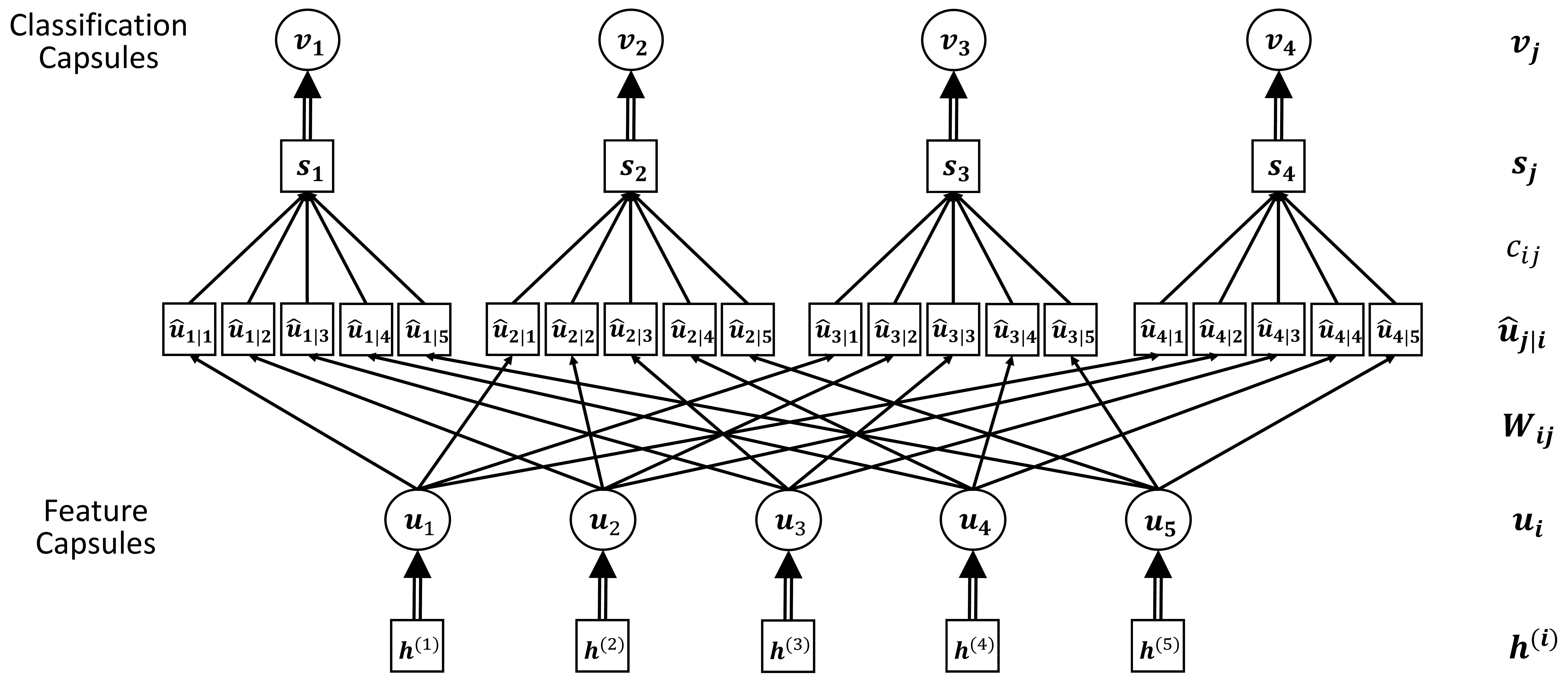}
\caption{The capsule layer where the double-line arrows indicate squashing functions (take four classification as an example).}
\label{Fig:CapsLayers}
\end{center}
\end{figure}

As illustrated in Fig.~\ref{Fig:CapsLayers}, the total input to a capsule $\bm{s}_j$ is a weighted sum over all
``prediction vectors'' $\bm{\widehat{u}}_{j|i}$ from the capsules in the layer below and is produced by multiplying the output $\bm{u}_i$ of a capsule in the layer below by a weight matrix $\bm{W}_{ij}$:
\begin{equation}
    \bm{s}_j=\sum_{i} c_{ij}\bm{\widehat{u}}_{j|i}
\end{equation}

\begin{equation}
    \bm{\widehat{u}}_{j|i} = \bm{W}_{ij} \bm{u}_i
\end{equation}
where $c_{ij}$ are coupling coefficients that are determined by an iterative dynamic routing algorithm with a given number of iterations $r$ (see \cite{Sabour2017Dynamic} for more details).

In training, a separate margin loss, $L_j$, for each classification capsule, $v_j$, is minimized:

\begin{equation}
    L_j = R_j~{\max(0, m^{+}-||\bm{v}_j||)}^2+\lambda (1-R_j)~{\max(0, ||\bm{v}_j||-m^{-})}^2
\end{equation}
where $R_j = 1$ iff the relation $j$ exists and $m^{+} = 0.9$, $m^{-} = 0.1$ and $\lambda =0.5$. The total loss is simply the sum of the losses of all classification capsules.

\section{Experimental Studies}
\label{Sec:Experimental Studies}

As to relation extraction, we perform experiments to first evaluate our recurrent capsule network, than evaluate the final severity classification of CAD.

\subsection{Dataset and evaluation metrics}
\label{SubSec:Dataset and Evaluation Metrics}

The dataset consists of coronary arteriography texts collected from Shanghai Shuguang Hospital. Shanghai Shuguang Hospital is located in Shanghai, which is one of the highest ranked hospitals in China. After CNER, we asked two students as a group under a doctor's guidance to manually annotate relations in these sentences and the final severity classification of CAD. Disagreements between the two annotators were resolved by the doctor. After the annotation, to make the dataset more suitable for our relation extraction task, we made several refinements as follows.
\begin{enumerate}
    \item We add direction to the relation names, such that ``r:percentage'' is splitted into two relations ``r:percentage(e1,e2)'' and ``r:percentage(e2,e1)'' except for ``no relation''. This leads to six relations in the dataset.
    \item We calculate the frequency of each relation with two directions separately. To better balance the relation dataset, $85\%$ ``no relation'' sentences are discarded.
    \item We performed standard random splitting on the relation dataset, with $70\%$ training and $30\%$ test sets.
\end{enumerate}

The statistical characteristics of the relation dataset are shown in Table~\ref{Tab:Dataset}, and the dataset of the final severity classification are shown in Table~\ref{Tab:DatasetClassification},

\begin{table}[!htbp]
\begin{center}
\caption{Statistical Characteristics of the Relation Dataset}
\label{Tab:Dataset}
\begin{tabular}{|c|c|c|c|}
\hline
Relation Name       & Training & Test & Total \\ \hline
r:modifier(e1,e2)   & 754      & 314  & 1,068  \\
r:negative(e2,e1)   & 296      & 110   & 406   \\
r:position(e1,e2)   & 257      & 132  & 389   \\
r:percentage(e1,e2) & 70       & 30   & 100   \\
r:percentage(e2,e1) & 176      & 80   & 256   \\
no relation         & 1,384     & 593  & 1,977  \\ \hline
Total               & 2,937     & 1,259 & 4,196 \\ \hline
\end{tabular}
\end{center}
\end{table}

\begin{table}[!htbp]
\begin{center}
\caption{Statistical Characteristics of the Final Severity Classification}
\label{Tab:DatasetClassification}
\begin{tabular}{|c|c|c|}
\hline
Severity Level      & Number  & Percentage  \\ \hline
Mild Stenosis       & 145     & 72.5\%    \\
Moderate Stenosis   & 45      & 22.5\%     \\
Severe Stenosis     & 10      & 5.0\%     \\ \hline
Total               & 200     & 100.0\%    \\ \hline
\end{tabular}
\end{center}
\end{table}

To evaluate the methods, we use the standard and widely-used performance metrics~\cite{liu2014strategy,Zhou2015CORRELATION}, i.e., precision (P), recall (R) and F$_1$-score (F$_1$) for relation extraction and the final severity classification of CAD. We also report the accuracy of the final severity classification of CAD.

\subsection{Implementation details}
\label{SubSec:Implementation Details}

After CNER, Chinese words are first segmented via Jieba Chinese segmentation module \footnote{\url{https://github.com/fxsjy/jieba}}, then pretrained 128-dimensional word vectors and entity type vectors in the embedding layer are obtained using the word2vec method~\cite{mikolov2013efficient} on both the training data and the test data, and they are updated during the training process. The size of each LSTM hidden states in bidirectional LSTMs (Bi-LSTMs) and unidirectional LSTMs (Uni-LSTMs) is set to 64 and 128, respectively. The dimension of the capsules is set to 64, and the iterative number $r$ of dynamic routing is set to 4. To minimize the margin loss, the whole network is trained by Adam optimization algorithm~\cite{kingma2014adam} with default parameter settings and the batch size is set to 128.

\subsection{Comparisons with baseline methods of relation extraction}
\label{SubSec:Comparisons With Baseline Methods}

We compare RCN models with five state-of-the-art methods. These five reference algorithms have been widely used for relation classification.
\begin{itemize}
    \item \textbf{CNN + MaxPooling}~\cite{nguyen2015relation}: It took words and their positions as input, and utilized a CNN with max-pooling to extract lexical and sentence level features. Finally, a softmax function is used to classify relations.
    \item \textbf{BiLSTM + MaxPooling}~\cite{zhang2015relation}: It sent words to a Bi-LSTM network with max-pooling, and used a softmax function to classify relations.
    \item \textbf{BiLSTM + Attention}~\cite{zhou2016attention}: It exploited an attention-based Bi-LSTM network to receive input words, and used a softmax function for relation classification.
    \item \textbf{CRNN + MaxPooling} and \textbf{CRNN + Attention}~\cite{raj2017learning}: They employed a CRNN architecture that combines RNNs and CNNs in sequence. \textbf{CRNN + MaxPooling} took words as input and utilized a max-pooling strategy along with a softmax function to classify relations. \textbf{CRNN + Attention} is an attention-based pooling technique.
\end{itemize}
Note that our model utilize entity type features which is not exploited in the baselines, we also report the results of the baselines with entity type features.

Table~\ref{Tab:CompareBaseline} shows the comparative results of our model and the baselines. First of all, we can observe that our model with entity type features outperforms these reference algorithms, with $95.59$\% in Precision, $97.45$\% in Recall and $96.51$\% in F$_1$-score. The improvements compared with the original baselines without entity type features are $1.5$, $1.5$, $1.05$, $0.45$ and $1.2$ points in Recall, $0.54$, $1.64$, $2.53$, $1.63$ and $1.77$ points in F$_1$-score, respectively. Without entity type features, our model also outperforms the reference algorithms without entity type features. Both the Recall and F$_1$-score of our model achieve the best ones, and the Precision of our model is just below the CNN + MaxPooling method, which use an extra position feature, while our model does not. Secondly, the entity type features proposed by us can help improve performance of the original baselines. The benefits in F$_1$-score brought by the entity type features are $0.27$, $0.87$, $2.28$, $0.69$ and $1.22$ points, respectively. However, the improved performance is still worse than our model. Thirdly, among the baselines, it is interesting to note that without entity type features, attention-based pooling technique performs worse than conventional max-pooling strategy, which has also been observed earlier by Sahu and Anand~\cite{sahu2017drug} and Raj et al.~\cite{raj2017learning}, while with entity type features, attention-based pooling technique performs better than conventional max-pooling strategy.

\begin{table*}[!htb]
\begin{center}
\caption{Comparative Results of Our RCN Model and Baseline Methods}
\label{Tab:CompareBaseline}
\begin{tabular}{|l|l|c|c|c|}
\hline
\multicolumn{1}{|c|}{Method}           & \multicolumn{1}{|c|}{Input}              & P     & R     & F$_1$ \\
\hline
\multirow{2}{*}{CNN + MaxPooling~\cite{nguyen2015relation}}    & Word + Position   & 95.54 & 96.40 & 95.97\\
& Word + Position + Entity Type* & \textbf{96.53} & 95.95 & 96.24 \\
\hline
\multirow{2}{*}{BiLSTM + MaxPooling~\cite{zhang2015relation}} & Word Only  & 94.45 & 94.30 & 94.87 \\
& Word + Entity Type*  & 95.52 & 95.95 & 95.74 \\
\hline
\multirow{2}{*}{BiLSTM + Attention~\cite{zhou2016attention}}  & Word Only  & 94.27 & 93.70 & 93.98 \\
& Word + Entity Type*  & 96.11 & 96.40 & 96.26 \\
\hline
\multirow{2}{*}{CRNN + MaxPooling~\cite{raj2017learning}}   & Word Only   & 93.97 & 95.80 & 94.88 \\
& Word + Entity Type*  & 94.18 & 97.00 & 95.57 \\
\hline
\multirow{2}{*}{CRNN + Attention~\cite{raj2017learning}}    & Word Only   & 93.70 & 95.80 & 94.74 \\
& Word + Entity Type*  & 95.68 & 96.25 & 95.96 \\
\hline
\multirow{2}{*}{\textbf{Our RCN model}} & Word Only & 95.14 & 96.85 & 95.99 \\
& Word + Entity Type* & 95.59 & \textbf{97.45} & \textbf{96.51} \\
\hline
\multicolumn{5}{l}{\begin{tabular}[c]{@{}l@{}}\begin{minipage}{5.9cm}\vspace{1mm}\tiny \item[*] The entity type features are proposed by us.\end{minipage} \end{tabular}}
\end{tabular}
\end{center}
\end{table*}

Furthermore, we compare class-wise performance of our RCN model with baseline methods. The comparative results are summarized in Table~\ref{Tab:CompareBaselineClass}. Firstly, we clearly observe that our model achieves No.1 in three relations (i.e. r:modifier[e1,e2], r:negative[e2,e1] and r:percentage[e2,e1]) and No.2 in one relation (i.e. r:percentage[e1,e2]) among all the five relations. The F$_1$-scores are higher than $90.00$ except for r:percentage(e1,e2) relation because of its low frequency (only 70 instances) in the training set. Secondly, we also obtain the same observation on the proposed entity type features as mentioned above that they can help improve performance of the original word embedding features.

\begin{table*}[!htb]
\begin{center}
\caption{Class-wise Performance (in Terms of F$_1$-score) of Our RCN Model and Baseline Methods}
\label{Tab:CompareBaselineClass}
\begin{tabular}{|l|l|c|c|c|c|c|}
\hline
\multicolumn{1}{|c|}{Method} & \multicolumn{1}{|c|}{Input}   & \begin{tabular}[c]{@{}c@{}}r:modifier\\(e1,e2)\end{tabular} & \begin{tabular}[c]{@{}c@{}}r:negative\\(e2,e1)\end{tabular} & \begin{tabular}[c]{@{}c@{}}r:position\\(e1,e2)\end{tabular} & \begin{tabular}[c]{@{}c@{}}r:percentage\\(e1,e2)\end{tabular} & \begin{tabular}[c]{@{}c@{}}r:percentage\\(e2,e1)\end{tabular} \\
\hline
\multirow{2}{*}{CNN + MaxPooling~\cite{nguyen2015relation}}    & Word + Position               & 96.33             & \textbf{99.54}    & 95.45       & 80.60          & 97.53\\
& Word + Position + Entity Type* & 96.28  & \textbf{99.54}  & \textbf{96.21}   & 80.00  & 98.77 \\
\hline
\multirow{2}{*}{BiLSTM + MaxPooling~\cite{zhang2015relation}}   & Word Only                             & 95.33             & 99.09             & 95.35             & 73.85               & 95.65\\
& Word + Entity Type*   & 96.66   & 99.09  & 94.62 & 81.16    & 96.20\\
\hline
\multirow{2}{*}{BiLSTM + Attention~\cite{zhou2016attention}}    & Word Only   & 94.87             & 99.09             & 93.44             & 73.85               & 93.17 \\
& Word + Entity Type*    & \textbf{97.15}             & 99.09             & 93.85             & \textbf{83.87}               & 98.14 \\
\hline
\multirow{2}{*}{CRNN + MaxPooling~\cite{raj2017learning}}  & Word Only    & 95.28             & 98.62             & 93.54             & 77.61               & 98.16 \\
& Word + Entity Type*  & 96.23      & \textbf{99.54}   & 94.70   & 80.00  & 96.34\\
\hline
\multirow{2}{*}{CRNN + Attention~\cite{raj2017learning}}   & Word Only  & 94.77 & 98.64 & 93.94 & 81.69               & 96.89 \\
& Word + Entity Type*  & 96.50  & \textbf{99.54}  & 94.62 & 80.60  & 98.16  \\
\hline
\multirow{2}{*}{\textbf{Our RCN model}} & Word Only & 96.35 & \textbf{99.54} & 93.08 & 76.47 & 98.14 \\
 & Word + Entity Type* & \textbf{97.15} & \textbf{99.54}  & 94.74             & 82.35   & \textbf{99.38}\\
\hline
\multicolumn{5}{l}{\begin{tabular}[c]{@{}l@{}}\begin{minipage}{5.9cm}\vspace{1mm}\tiny \item[*] The entity type features are proposed by us.\end{minipage}\end{tabular}}
\end{tabular}
\vspace{-3mm}
\end{center}
\end{table*}

\subsection{Evaluation of severity classification of CAD}
\label{SubSec:Evaluation of severity classification}

To evaluate the effectiveness of our proposed severity classification method, we randomly select 200 coronary arteriography texts for evaluation. The results are shown in Table~\ref{Tab:FinalClassification}. First of all, our method obtains an overall Accuracy of 97.00\%. Only six texts are classified into the wrong level. Secondly, most of the coronary arteriography texts (72.5\%) belongs to mild stenosis in practice. Our method achieves an relative high performance in terms of Precision (100\%), Recall (98.62\%) and F$_1$-score (99.31\%). Thirdly, our method is a little confused with moderate stenosis and severe stenosis. The precision, recall and F$_1$-score of moderate stenosis are all 93.33\%. It is still acceptable. As to severe stenosis, which appears rarely (5\%) in practice, though the precision is merely 75.00\%, but the recall is 90\%. That is to say, only one text of severe stenosis is not recognized by our method.

\begin{table}[!htb]
\begin{center}
\caption{Performance of Our Automatic Severity Classification Method}
\label{Tab:FinalClassification}
\begin{tabular}{|c|c|c|c|}
\hline
                  & P        & R      & F$_1$      \\ \hline
Mild Stenosis     & 100.00   & 98.62  & 99.31  \\
Moderate Stenosis & 93.33    & 93.33  & 93.33  \\
Severe Stenosis   & 75.00    & 90.00  & 81.82  \\ \hline
Overall Accuracy  & \multicolumn{3}{c|}{97.00} \\ \hline
\end{tabular}
\end{center}
\end{table}

\section{Analysis and Discussion}
\label{Sec:Analysis and Discussion}

In this section, we conduct two groups of experiments to respectively investigate the effect of the input features with different routing iterations, and the interest of the Uni/Bi-LSTMs and the capsule layer.

\subsection{Effect of the input features with different routing iterations}
\label{SubSec:Effect of the Input Features With Different Routing Iterations}

To study the effect of the input features and the routing iterations in our model, we experimentally compare the performance between different input features with different routing iterations. As shown in Table~\ref{Tab:DifferentFeatures}, our model achieves the best performance when adopting words and their entity types as input with 4 routing iterations. The Precision, Recall and F$_1$-score are $95.59$\%, $97.45$\% and $96.51$\%, respectively. Comparing the model inputs, we can observe that the performance of our model with entity type features is better than that without entity type features. The benefits are $0.66$ in Precision, $0.63$ in Recall and $0.65$ in F$_1$-score on average. Comparing the five lines in the table, we can observe that whatever inputs are used, the performance (in terms of F$_1$-score) first grows then drops down with the increase of iterative number $r$. When words are only used, our model achieves its best performance with $r = 2$. And when words and entity types are both used, our model achieves its best performance with $r = 4$.

\begin{table}[!htb]
\begin{center}
\caption{Comparisons between Different Input Features with Different Routing Iterations}
\label{Tab:DifferentFeatures}
\begin{tabular}{|c|c|c|c|c|c|c|}
\hline
\multirow{2}{*}{$r$} & \multicolumn{3}{c|}{Word Only} & \multicolumn{3}{c|}{Word + Entity Type} \\ \cline{2-7}
& P       & R        & F$_1$       & P          & R          & F$_1$ \\
\hline
1  & 94.53    & 95.80    & 95.16    & 94.97       & 96.25       & 95.61       \\
2  & \textbf{95.14}    & 96.85    & \textbf{95.99}    & 94.74       & 97.30       & 96.01 \\
3  & 94.69    & 96.25    & 95.46    & 95.15       & 97.15       & 96.14       \\
4  & 94.95    & 95.80    & 95.37    & \textbf{95.59}      & \textbf{97.45}  & \textbf{96.51} \\
5  & 93.25    & \textbf{97.30}    & 95.23    & 95.43       & 97.00       & 96.21 \\
\hline
\end{tabular}
\end{center}
\end{table}

\subsection{Interest of Uni/Bi-LSTMs and capsules}
\label{SubSec:Interest of Uni/Bi-LSTMs and Capsules}

To analyze the interest of Uni/Bi-LSTMs and capsules, we compare our model with that all using Bi-LSTMs or replacing the capsule layer by a fully-connected layer with a softmax function. The results are illustrated in Table~\ref{Tab:DifferentLayers}. From the table, we can observe that the F$_1$-score of our model is higher than that all using Bi-LSTMs by $0.29\%$, and higher than that using a softmax layer by $0.35$ point. It indicates the interest of the proposed Uni/Bi-LSTMs and capsules.

\begin{table}[!htb]
\begin{center}
\caption{Comparison of the Interest of Uni/Bi-LSTMs and Capsules}
\label{Tab:DifferentLayers}
\begin{tabular}{|c|c|c|c|}
\hline
              & P       & R        & F$_1$  \\
\hline
All Bi-LSTMs  & 94.76    & $\textbf{97.60}$    & 96.16 \\
Softmax layer & 95.16    & 97.30    & 96.22          \\
$\textbf{Uni/Bi-LSTMs \& Capsule layer}$      & $\textbf{95.59}$    & 97.45    & $\textbf{96.51}$ \\
\hline
\end{tabular}
\end{center}
\end{table}

\section{Conclusions and Future Works}
\label{Sec:Conclusions and Future Works}

In this paper, we present an effective method for the severity classification of coronary artery disease in EHRs. In our method, a recurrent capsule network model is employed to extract semantic relations in coronary arteriography texts. Specifically, words and their corresponding entity type features are first transferred into embedding vectors, then fed into a recurrent layer to capture high-level features. Finally, a capsule layer is used for relation classification.
Experimental results on the corpus collected from Shanghai Shuguang Hospital shows that our RCN model achieved an F$_1$-score of $96.41\%$ in relation extraction and an Accuracy of 97.0\% in the final severity classification of CAD. In future, we plan to use our recurrent capsule network model to solve other NLP tasks.

\section*{Acknowledgment}

We would like to thank the reviewers for their useful remarks and suggestions. This work was supported by the National Key R\&D Program of China for ``Precision Medical Research'' (No. 2018YFC0910500) and the National Natural Science Foundation of China (No. 61772201), and National Major Scientific and Technological Special Project for ``Significant New Drugs Development'' (No. 2018ZX09201008).

\bibliographystyle{IEEEtran}
\bibliography{IEEEexample}

\end{CJK*}
\end{document}